\documentclass{article}

\usepackage[english]{babel}
\usepackage[letterpaper,top=2cm,bottom=2cm,left=3cm,right=3cm,marginparwidth=1.75cm]{geometry}
\usepackage[square,numbers,sort&compress]{natbib}
\usepackage[font=small,labelfont=bf]{caption}

\usepackage{amsmath}
\usepackage{graphicx}
\usepackage[colorlinks=true, allcolors=blue]{hyperref}
\usepackage{slashbox}
\usepackage{xcolor}

\usepackage{color-edits}
\addauthor{JWV}{red}

\newcommand{\answerYes}[1][]{\textcolor{blue}{[Yes] #1}}
\newcommand{\answerNo}[1][]{\textcolor{orange}{[No] #1}}
\newcommand{\answerNA}[1][]{\textcolor{gray}{[N/A] #1}}
\newcommand{\answerTODO}[1][]{\textcolor{red}{\bf [TODO]}}

\title{Has the Machine Learning Review Process Become More Arbitrary as the Field Has Grown? \\
The NeurIPS 2021 Consistency Experiment\footnote{This report is an expanded version of an earlier blog post from December 2021\cite{consitencyblogpost}.}}
\author{Alina Beygelzimer, Yann N. Dauphin, Percy Liang, and Jennifer Wortman Vaughan}

\begin{document}
\maketitle

\begin{abstract}
We present the NeurIPS 2021 consistency experiment, a larger-scale variant of the 2014 NeurIPS experiment~\cite{neurips14experiment, CL21} in which 10\% of conference submissions were reviewed by two independent committees to quantify the randomness in the review process.  We observe that the two committees disagree on their accept/reject recommendations for 23\% of the papers and that, consistent with the results from 2014, approximately half of the list of accepted papers would change if the review process were randomly rerun. Our analysis suggests that making the conference more selective would increase the arbitrariness of the process. Taken together with previous research, our results highlight the inherent difficulty of objectively measuring the quality of research, and suggest that authors should not be excessively discouraged by rejected work.
\end{abstract}

\section{Introduction}

Across academic disciplines, peer review is used as a mechanism to vet the quality of research and identify interesting work.  However, like other human judgments~\cite{kahneman2021noise}, reviews are noisy, and studies across fields have shown that reviewers often disagree with one another~\cite{cicchetti1991reliability,bornmann2010reliability,obrecht2007examining, fogelholm2012panel,pier2017your}.  Within the field of machine learning, Corinna Cortes and Neil Lawrence, the 2014 program chairs of the top-tier conference Neural Information Processing Systems (NeurIPS), ran an experiment in which 10\% of NeurIPS submissions were reviewed by two independent program committees to quantify the randomness in the review process~\cite{neurips14experiment, CL21}, a type of noise audit~\cite{kahneman2021noise}.  They found that there was a high degree of inconsistency between reviewers.  One particularly salient result implied that if the review process had been independently rerun with a different assignment of reviewers to papers, approximately half of the accepted papers would have been rejected.

Since the time of that experiment, the impact and ubiquity of machine learning has only grown. The number of annual submissions to NeurIPS has increased more than fivefold, with more than 9,000 papers submitted in each of 2020, 2021, and 2022. This growth has led to rapidly expanding the pool of reviewers, with upwards of 9,000 program committee members playing a role in the review process in a given year.  

While serving as NeurIPS program chairs in 2021, we wanted to measure the extent to which the consistency in reviewers' decisions has changed as the conference has grown.  We therefore ran a variant of the 2014 consistency experiment during the 2021 review process.  As in the original experiment, we duplicated 10\% of submitted papers, assigned the two copies of each paper to two independent committees for review, and compared their recommendations.

Echoing prior work, our results suggest that the review process contains a high level of noise and subjectivity.  Consistent with the findings from 2014, we observed that about half of the list of accepted papers would have changed if we independently reran the review process.  Our analyses suggest that paper outcomes would only grow more arbitrary if the conference were made more selective, but that the acceptance rate could be increased without significantly impacting how arbitrary decisions are.  Taken together with results from a complementary study that showed high levels of disagreement between authors and reviewers and even between pairs of co-authors~\cite{perception22}, these results indicate an inherent difficulty in objectively measuring the merits of research, raising important issues for the research community to grapple with.

In this paper, we provide some background on the NeurIPS 2021 review process, detail the way in which the consistency experiment was implemented, and present an analysis of the results. We conclude with a discussion of the implications and limitations of the experiment.

\section{Background on the NeurIPS 2021 Review Process}
\label{sec:background}

Before describing the details of the experiment, we give some background on the way the NeurIPS review process was run in 2021.
The review process took place between May 28, 2021, when submissions were due, and September 28, 2021, when authors were notified of accept/reject decisions.  For the first time, the entire review process was conducted on OpenReview, a flexible and customizable peer review platform.  As in previous years, the review process was confidential; submissions under review were visible only to assigned program committee members. After the review process concluded, reviews for accepted papers were released publicly.\footnote{\url{https://openreview.net/group?id=NeurIPS.cc/2021/Conference}}  Authors of rejected papers were given the opportunity to opt in to have their reviews publicly released, but they were kept private by default.  The review process was double-blind, meaning that authors were required to anonymize their submissions and were not allowed to include any information that might reveal their identity to reviewers.  One new feature added in 2021 was that authors were asked to include in their submission a checklist designed to encourage best practices for responsible machine learning research, including issues of reproducibility, transparency, research ethics, and societal impact~\cite{checklistblogpost}; in the interest of promoting more responsible research practices, a completed paper checklist for this report is included in Appendix~\ref{app:checklist}. 

The program committee was made up of more than 9,000 program committee members playing different roles.
\emph{Reviewers} were responsible for reading and evaluating an assigned set of submissions and participating in discussions on each paper. 
\emph{Area chairs (ACs)} were responsible for recommending reviewers for submissions, ensuring that all submissions received high-quality reviews, facilitating discussions among reviewers, writing meta-reviews, evaluating the quality of reviews, and making decision recommendations. 
\emph{Senior area chairs (SACs)} each oversaw the work of a small number of ACs, making sure that the review process went smoothly. SACs were also responsible for helping ACs find expert reviewers, calibrating decisions across ACs, discussing borderline papers, and helping the program chairs make final decisions. 
\emph{Ethics reviewers} provided additional reviews for submissions flagged for potential ethical concerns. Their comments were intended to inform reviewer deliberations. 
In total there were more than 8,000 reviewers, 708 ACs, 95 SACs, and 105 ethics reviewers, plus the 4 program chairs. Program committee members were not compensated for their service.

The content of the review forms used by reviewers and ACs is included in Appendix~\ref{app:forms}.
Initial reviews were released to authors on August 3, 2021, at which point 99.7\% of submissions had at least three reviews available.  Authors were given one week to submit a response to the reviews before the reviewer discussion period began on August 10.  To minimize the chance of misunderstandings, there was also an opportunity for additional rolling discussion between authors and the program committee after the initial response period. If new reviews were added (including ethics reviews), authors had the opportunity to respond to those during the rolling discussion.

In total, the conference received 9,122 submissions, of which 2,334 were accepted, for an acceptance rate of 25.6\%.  Out of 2,334 accepted papers, 55 papers were accepted as oral presentations, 260 as spotlights, and the remaining 2,019 as posters only.  Decisions to accept papers as oral presentations or spotlights were made jointly between SACs and the program chairs.

In addition to the experiment described in this paper, we ran a survey to understand authors' perceptions about the quality of their submitted papers as well as their perceptions of the peer-review process.  Specifically, we asked submitting authors to report their predicted probability of acceptance for each of their papers, their perceived ranking of their own papers based on scientific contribution, and the change in their perception of their own papers after seeing the reviews.  We found that authors overestimate the probability their papers will be accepted by roughly a factor of three, with a median prediction of about 70\% when the acceptance rate was 25.8\%.  In cases where an author had one paper accepted and another rejected, authors ranked the rejected paper as higher quality about a third of the time. Surprisingly, when co-authors ranked their jointly authored papers, they also disagreed with each other about the relative merits of their paper about a third of the time. As discussed in Section~\ref{sec:discussion}, this suggests an inherent difficulty, or perhaps impossibility, of objectively quantifying the merits of a paper.  The results of this experiment are explored in a separate report~\cite{perception22}.

\section{Methods}
\label{sec:methods}

During the assignment phase of the review process, we chose 10\% of papers uniformly at random to duplicate.  We'll refer to these as the \emph{duplicated papers}.  We assigned two ACs and twice the usual number of reviewers to these papers.  With the help of the team at OpenReview, we then created a copy of each of these papers and split the ACs and reviewers at random between the two copies.  We made sure that the two ACs were assigned to two different SACs so that no SAC handled both copies of the same paper.  Any newly invited reviewer for one copy was automatically added as a conflict for the other copy.  We'll refer to the SAC, AC, and reviewers assigned to the same copy as the copy's \emph{committee}.

We note that this implementation is slightly different from what was done in 2014.  In 2014, the entire program committee was randomly split into two committees for the 10\% of the papers in the experiment.  One of the program committees was then marked as conflicted with the originals and the other was marked as conflicted with the copies of papers in the experiment.  To allow for more flexibility in the assignment of reviewers, we instead effectively created a different split for each paper in the experiment.

To mitigate any influence the experiment might have on the program committee's behavior, the duplicated papers' committees were not told about the experiment and were not aware the paper had been duplicated.  The OpenReview team reused paper IDs of withdrawn abstract submissions for the duplicated papers to avoid revealing which papers were duplicated.  The authors of duplicated papers were notified of the experiment right before initial reviews were released and instructed to respond to each set of reviews independently.  They were also asked to keep the experiment confidential. The email that was sent to authors is included in Appendix~\ref{app:authoremail}.

As in 2014, duplicated papers were accepted if at least one of the two copies was recommended for acceptance and no ``fatal flaw'' was found.  This resulted in 92 accepted papers that would not have been accepted had we not run the experiment, increasing the overall conference acceptance rate from 24.6\% to 25.6\%.  Four papers that were accepted by one committee (one as a spotlight, three as posters) were ultimately rejected due to what was considered a fatal flaw.  In an additional two cases, the committees for the two papers disagreed about whether a flaw was ``fatal.''  In these cases, the papers were conditionally accepted with conditions determined jointly by the two committees; both were ultimately accepted.

At the time initial reviews were released, 8,765 of the original 9,122 submitted papers were still under review, and 882 of these were duplicated papers.  This set of 882 papers is what we consider in the analyses in Section~\ref{sec:results}.  This means we do not include duplicated papers that were desk rejected for violations of the call for papers or withdrawn by the authors before initial reviews were released; reviewer scores and acceptance decisions were not available for these papers and the authors of these papers never learned that they were part of the experiment.  We do include the 118 duplicated papers that were withdrawn by the authors after initial reviews were released since authors were more likely to withdraw a paper after seeing negative reviews. We note that the withdrawal rate after seeing initial reviews was 45\% higher for papers not in the experiment compared with duplicated papers, which we suspect is because authors of duplicated papers had two shots at acceptance.

This experiment was independently reviewed and approved by an Institutional Review Board (IRB) which determined the risks to participants were no greater than those experienced by participating in the normal review process.  Since the data collected (i.e., submitted papers and their authors, assignments of reviewers to papers, content of the reviews, discussion comments, and so on) contains personally identifiable information, we agreed that only the 2021 program chairs, workflow manager, and the OpenReview staff would access this data. However, for all duplicated papers that were accepted (and any rejected duplicated papers that opted in to make their reviews public), both sets of reviews are publicly available on OpenReview.  The conference call for papers\footnote{\url{https://neurips.cc/Conferences/2021/CallForPapers}} included a statement alerting authors that ``as in past years, the program chairs will be measuring the quality and effectiveness of the review process via randomized controlled experiments.''

\section{Results}
\label{sec:results}

\begin{figure}[t!]
\centering
\includegraphics[width=0.5\textwidth]{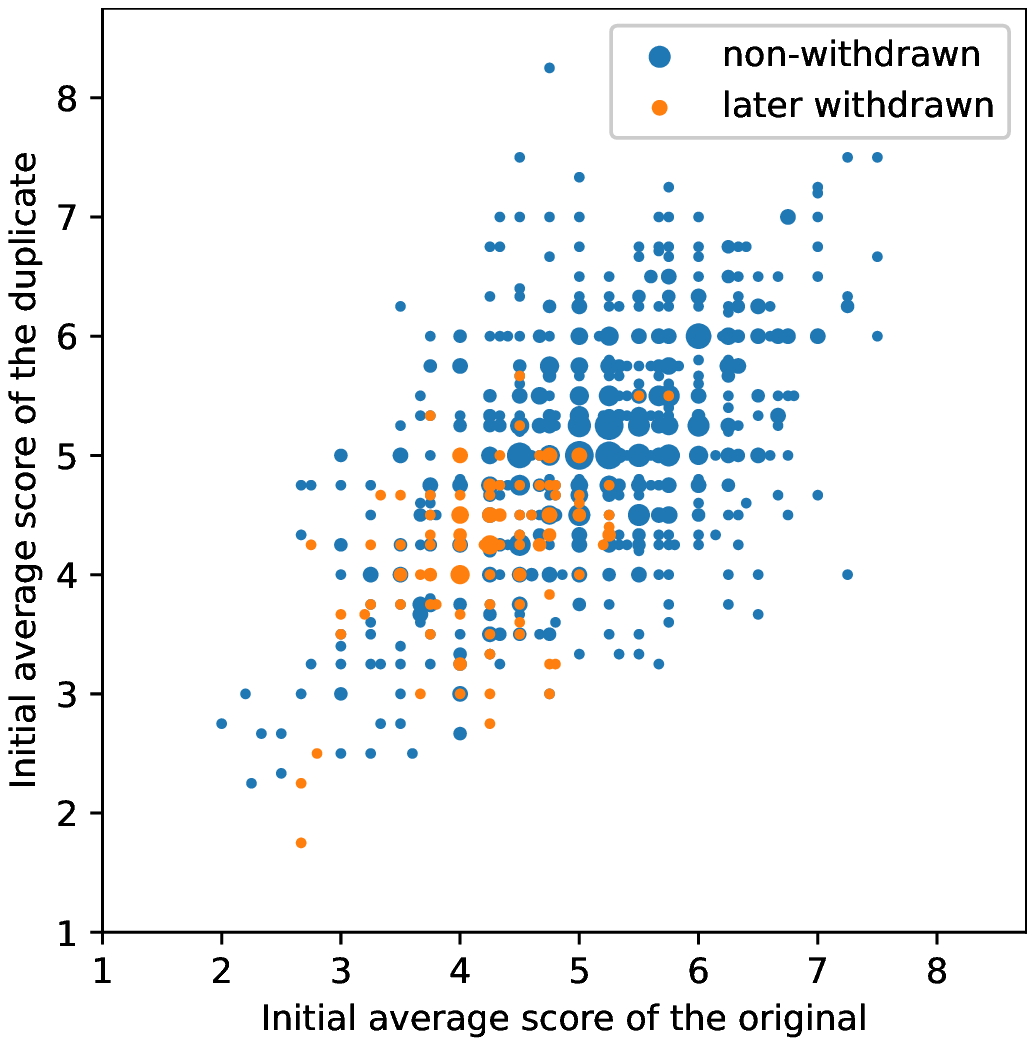}
\includegraphics[width=0.5\textwidth]{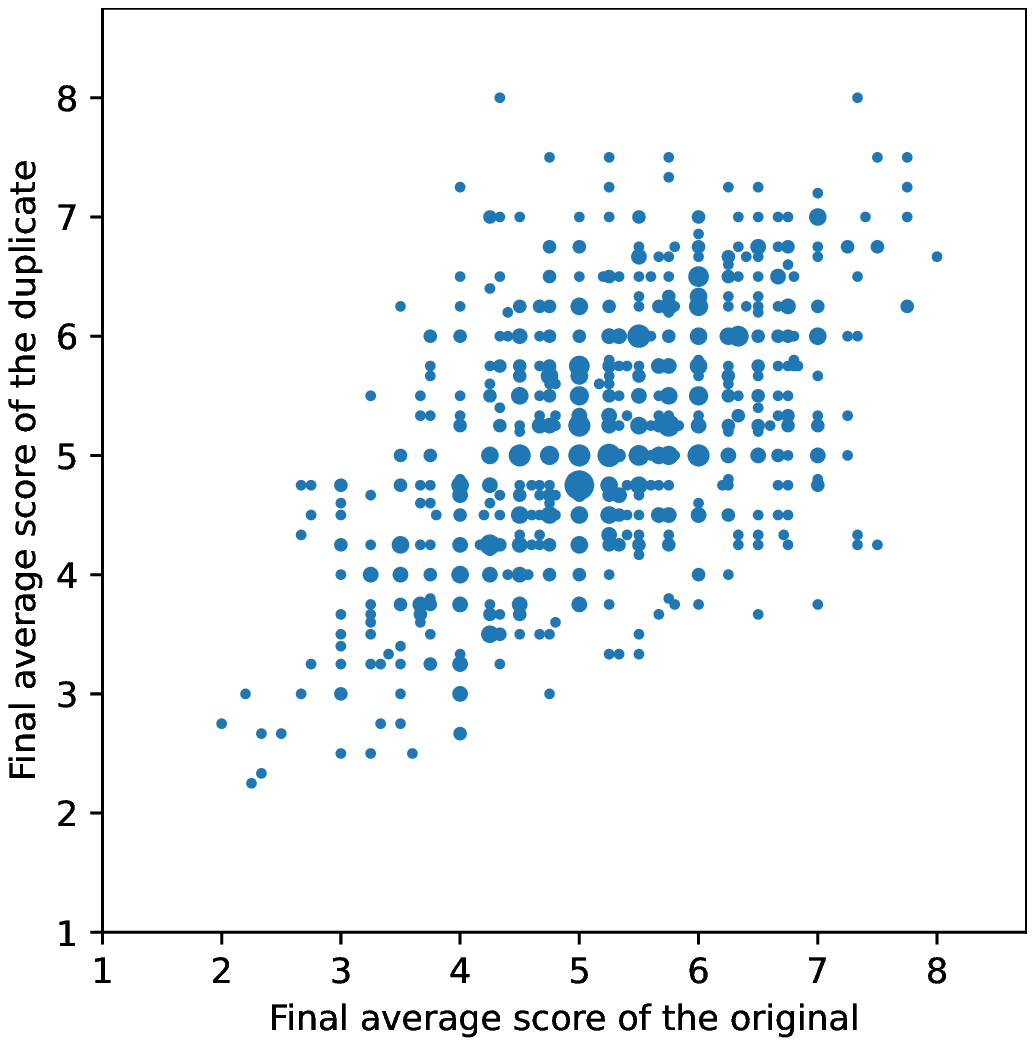}
\caption{\label{fig:scatterplot} A scatter plot showing average scores of the two committees at the time that initial reviews were released (top) and at the time final scores were released (bottom) for each paper in the experiment. The area of each circle is linear in the number of papers that it represents. The Pearson correlation coefficient is 0.575 for initial scores and 0.586 for final scores.}
\end{figure}

Table~\ref{tab:decisions} summarizes the recommendations for the 882 duplicated papers still under review when initial reviews were released.  We discuss several interpretations of these results below.  Figure~\ref{fig:scatterplot} shows the correlation between average initial reviewer scores (i.e., overall scores or ``ratings'' at the time initial reviews were released) from the two committees assigned to each paper, and the same for average final reviewer scores.




\begin{table}[ht!]
\centering
\begin{tabular}{|l||*{7}{c|}}\hline
\backslashbox{Original}{Copy}
& Oral & Spotlight & Poster
& Reject & Withdrawn \\\hline\hline
Oral & 0 & 0 & 4 & 0 & 0\\\hline
Spotlight & 0 & 3 & 9 & 13 & 0\\\hline
Poster & 2 & 7 & 74 & 94 & 0\\\hline
Reject  & 0 & 13 & 83 & 462 & 0\\\hline
Withdrawn  & 0 & 0 & 0 & 0 & 118\\\hline
\end{tabular}
\caption{Summary of the recommendations for the 882 duplicated papers that were still under review on August 3, 2021, when initial reviews were released to authors.\label{tab:decisions}}
\end{table}

\subsection{Inconsistent Decisions}
\label{sec:id}

There are a few ways to think about the results.  First, we can measure the fraction of inconsistent accept/reject recommendations---the fraction of duplicated papers that were accepted by only one of the two committees, grouping together oral presentations, spotlights, and posters. The number of papers with such inconsistent recommendations was 203 out of 882, or 23.0\%.

To put this number in context, we need a baseline.  There were 206 papers accepted in the original set and 195 papers accepted in the duplicate set, for an average acceptance rate of 22.7\%.  If acceptance recommendations were made at random with a 0.227 chance of accepting each paper, we would expect the fraction of inconsistent recommendations to be 35.1\%.  While the fraction of inconsistent recommendations is closer to the random baseline than it is to 0, many of these papers could genuinely have gone either way.  When ACs entered recommendations, they were asked to note whether they were sure or whether the paper could be bumped up or down.  (See the full meta-review form in Appendix~\ref{app:forms}.)  If we treat the pair ``poster accept that can be bumped down'' and ``reject'' as consistent, and do the same for the pair ``reject that can be bumped up'' and ``poster accept that should not be bumped up to spotlight,'' then the fraction of inconsistent recommendations drops to only 16.0\%.

\begin{figure}[t!]
\centering
\includegraphics[width=0.7\textwidth]{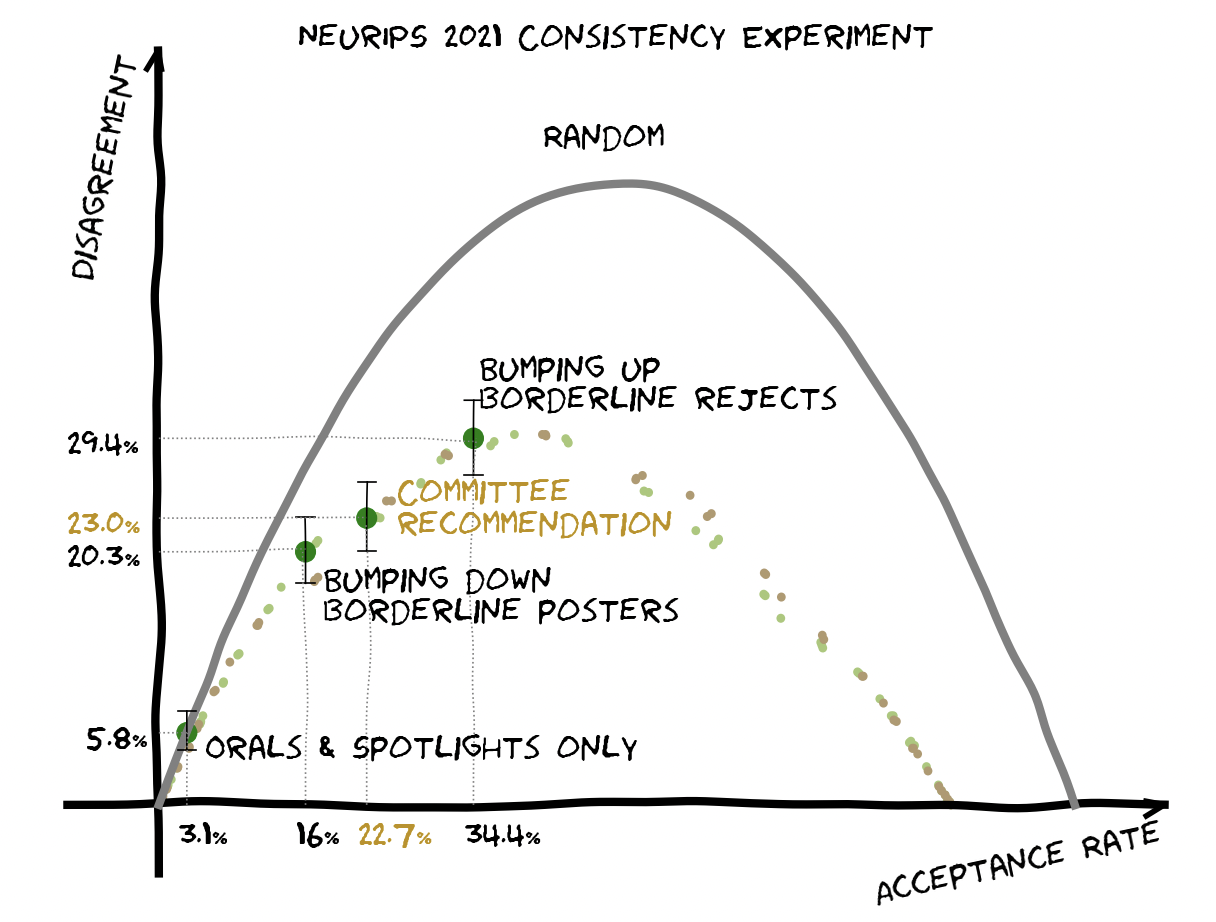}
\caption{\label{fig:dis_vs_accept}  Acceptance rates and amount of disagreement between the two committees using different acceptance thresholds. The gray curve shows the random baseline: for each potential acceptance rate, the fraction of papers for which there would be disagreement between two committees if their recommendations were made at random. 
Small green (respectively, brown) dots show, for each acceptance rate, the level of disagreement there would have been between the two committees if the papers with the highest average final (respectively, initial) reviewer scores were accepted. Error bars depict Wilson's confidence intervals.}
\end{figure}

We can see how the fraction of inconsistent recommendations would have changed if we shifted the acceptance threshold in different ways.  For example, if the conference were so selective as to accept only those papers that were assigned orals and spotlights, the committees would have accepted 29 and 25 of the duplicated papers respectively, agreeing on only 3 papers.  To visualize the impact of shifting the acceptance threshold, the small green dots in Figure~\ref{fig:dis_vs_accept} show, for each acceptance rate $x$, the level of disagreement there would have been between the two committees if the papers with the highest $x$\% of average final reviewer scores were accepted; the brown dots show the same for average initial reviewer scores (i.e., average scores at the time reviews were first released to authors).
The gray curve in Figure~\ref{fig:dis_vs_accept} extends the random baseline described above to other acceptance rates.  Points on the gray curve correspond to the expected fraction of inconsistent recommendations if both committees were making acceptance recommendations at random with the corresponding acceptance probability.  
We have additionally included the following points on the plot:
\begin{itemize}
\item Accepting only papers recommended as orals or spotlights leads to a 5.8\% disagreement rate, but this is only an 8\% relative improvement over the random baseline at the corresponding acceptance rate of 3.1\%.
\item Bumping down all posters marked as candidates to bump down leads to a 20.3\% disagreement rate, a 25\% relative improvement over the random baseline at the acceptance rate of 16.0\%.
\item As mentioned above, the recommendations made by the NeurIPS 2021 committees led to a disagreement rate of 23.0\%, a 35\% relative improvement over the random baseline at the acceptance rate of 22.7\%.
\item Bumping up all rejects that were marked as candidates for being bumped up leads to a 29.4\% disagreement rate, which is also a 35\% relative improvement over the random baseline at the corresponding acceptance rate of 34.4\%.
\end{itemize}

For comparison, in 2014, of the 166 papers that were duplicated, the two committees disagreed on 43 (25.9\%).  The acceptance rate was 25\% for duplicated papers---a bit higher than the overall 2014 acceptance rate.  The random baseline for this acceptance rate is 37.5\% disagreement, so this is a 31\% relative improvement.  Given the small sample size in the 2014 experiment, there is a fairly large confidence interval around these numbers, but the 2021 results appear to be in line with those from 2014, with no obvious increase or decrease in inconsistency.

\subsection{Accept Precision}
\label{sec:ap}

Another way of measuring disagreement is to look at the fraction of accepted papers that would have changed if we reran the review process.  This is also the probability that a randomly chosen accepted paper would have been rejected if it were re-reviewed, previously discussed as the (complement to 1 of) ``accept precision'' or ``arbitrariness'' in the context of the 2014 experiment~\cite{lawrencetalk,cacmblog,mrtzblog}.

\begin{figure}[t!]
\centering
\includegraphics[width=0.7\textwidth]{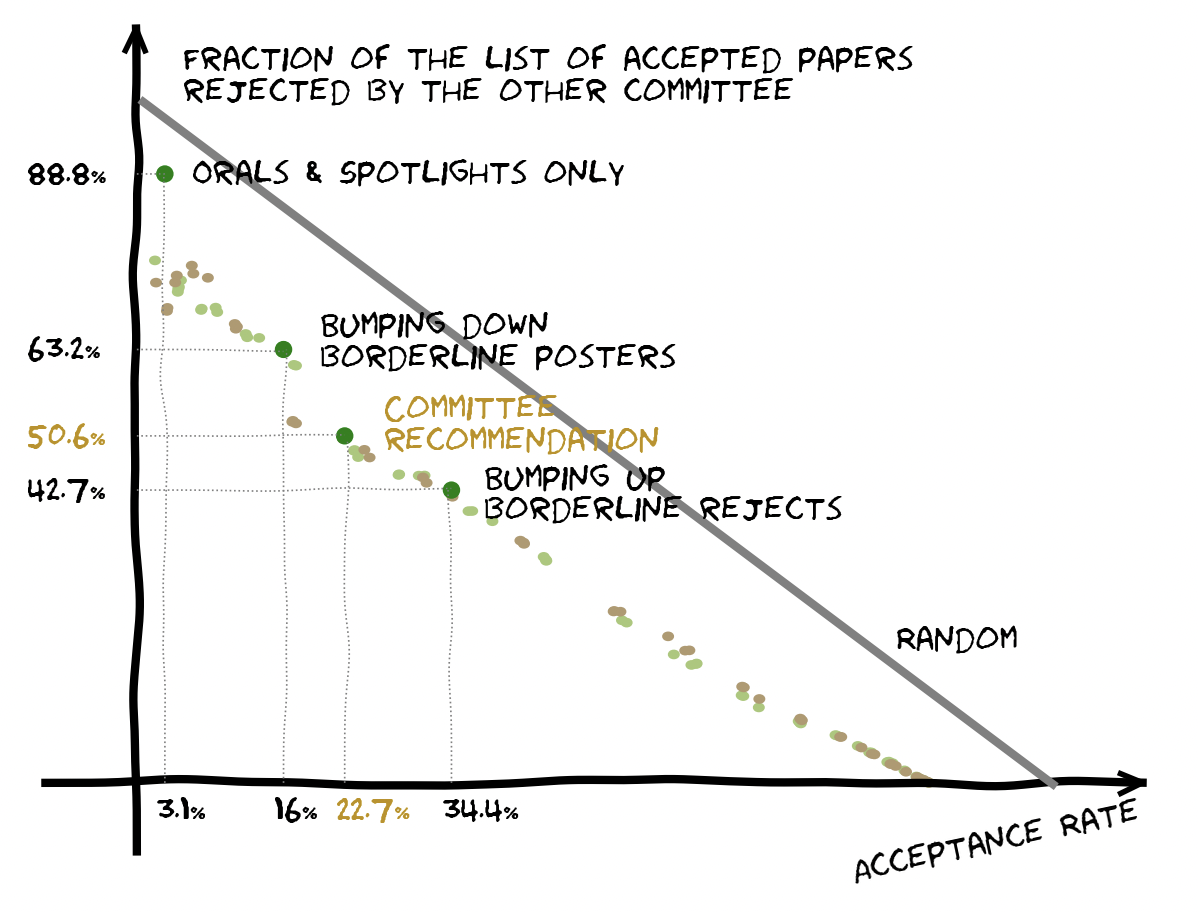}
\caption{\label{fig:fracchanged_vs_accept} The fraction of accepted papers that would be rejected by the other committee, using different acceptance thresholds.  The gray line shows the random baseline: for each potential acceptance rate, the fraction of accepted papers that would be rejected by the other committee if recommendations were made at random. Small green (respectively, brown) dots show, for each acceptance rate, the fraction of accepted papers that would be rejected by the other committee if the papers with the highest average final (respectively, initial) reviewer scores were accepted.}
\end{figure}

In 2014, 49.5\% of the papers accepted by the first committee were rejected by the second (with a fairly wide confidence interval as the experiment included only 116 papers).  This year, this number was 50.6\% across the two committees (51.9\% for one committee and 49.2\% for the other).  This is a 35\% improvement over the random baseline, i.e., the fraction of accepted papers that would be rejected by the other committee if recommendations were made at random with the same acceptance rate.  

Figure~\ref{fig:fracchanged_vs_accept} shows how this number would change if different acceptance thresholds were used.  Similar to what we observed for inconsistent recommendations, choosing a more selective acceptance rate leads to lower accept precision:
\begin{itemize}
\item Accepting only papers recommended as orals or spotlights yields only an 8\% relative improvement over the random baseline, with 88.8\% of accepted papers rejected by the other committee.
\item Bumping down all posters marked as candidates for being bumped down leads to a 25\% relative improvement over the random baseline, with 63.2\% of accepted papers rejected by the other committee.
\item As noted above, the recommendations made by NeurIPS 2021 committees led to a 35\% relative improvement over the random baseline, with 50.6\% of accepted papers rejected by the other committee.
\item Bumping up all rejects that were marked as candidates for being bumped up also yields a 35\% relative improvement over the random baseline, with 42.7\% of accepted papers rejected by the other committee.
\end{itemize}

We note that there is especially high arbitrariness on which papers are selected for orals and spotlights. These recommendations are made by the ACs, SACs, and program chairs taking into account not only raw reviewer scores but other factors, such as interest to a broad audience.  Choosing papers to highlight with oral presentations and spotlights may be a more subjective task than simply assigning a score to a paper. 

We can also look at the probability that a randomly chosen rejected paper would have been accepted if it were re-reviewed.  This number was 14.9\% this year, compared to 17.5\% in 2014.

There were 24 duplicated papers where each committee was unanimous internally about the accept/reject decision yet the two committees disagreed with each other on what that decision should be.

\subsection{Ethics Flags}

As discussed in a NeurIPS blog post reflecting on the 2021 ethics review process~\cite{ethicsblog}, one of the biggest challenges in implementing ethics review at NeurIPS is the uncertainty that reviewers face around which papers to flag. This uncertainty adds noise into the process and  leads to inconsistency in which papers receive ethics reviews.  Examining how reviewers flagged duplicated papers for ethics review emphasizes this point.  There were 23 papers flagged by one committee and 22 papers flagged by the other, but the overlap between these two sets was only 3 papers---a little over 13\%.

\subsection{Feedback from ACs and SACs}

After the review period ended and decisions were released, we gave ACs and SACs who were assigned to duplicated papers for which there had been disagreement between committees access to the reviews and discussion for the papers' other copies.  We asked them to complete a brief survey to provide feedback on the experiment.  Of the 203 papers that were recommended for acceptance by only one committee, we received feedback on 99.  Unfortunately, we received feedback from both committees for only 18 papers, which limits the scope of our analysis.

Based on this feedback, the vast majority of cases fell into one of three categories.  First, there is what one AC called ``noise on the decision frontier.''  In such cases, there was no real disagreement, but one committee may have been feeling a bit more generous or more excited about the work and willing to overlook the paper's limitations.  Indeed, 48\% of multiple-choice responses were ``This was a borderline paper that could have gone either way; there was no real disagreement between the committees.''

Second, there were genuine disagreements about the value of the contribution or the severity of limitations.  We saw a spectrum here ranging from basically borderline cases to a few more difficult cases in which expert reviewers disagreed.  In some of these cases, there was also disagreement within committees.

Third were cases in which one committee found a significant issue that the other did not.  Such issues included, for example, close prior work, incorrect proofs, and methodological flaws.

45\% of responses were ``I still stand by our committee's decision,'' while the remaining 7\% were ``I believe the other committee made the right decision.''  We can only speculate about why this may be the case.  Part of this could be that, once formed, opinions are hard to change.  Part of it is that many of these papers are borderline, and different borderline papers just fundamentally appeal to different people.  Part of it could also be selection bias; the ACs and SACs who took the time to respond to our survey may have been more diligent and involved during the review process as well, leading to better decisions.

\section{Discussion}
\label{sec:discussion}

A common complaint among members of the machine learning research community is that the NeurIPS review process has become noisier as the conference has grown. In contrast, our analysis shows no evidence that the review process has become noisier with increasing scale. All results appear to be in line with those from the 2014 consistency experiment. On some level, we can view this as good news.

Still, there is significant noise in the review process.  This may be due in part to fundamental limits on how consistent reviews can be. In a recent analysis revisiting the results of the original 2014 consistency experiment, \citet{CL21} argue that about 50\% of the variance in recommendations can be attributed to subjective opinions.  In a parallel and complementary study that we ran on NeurIPS 2021 authors' perceptions of the quality of their own work~\cite{perception22}, we found high levels of disagreement on paper quality between authors and reviewers, but also, perhaps more surprisingly, high disagreement between co-authors on the relative quality of their own co-authored submissions. This suggests a fundamental difficulty in objectively ranking papers.  Subjectivity and noise are common in other forms of decision making as well~\cite{kahneman2021noise} and while it may be possible to implement interventions to reduce the level of noise, we will never be able to avoid some level of randomness in the process.

In light of this, we would encourage authors to avoid excessive discouragement from rejections. The NeurIPS 2014 program chairs found that more than a third of papers that were rejected from NeurIPS were ultimately published in other high-quality venues and we expect the same of many papers rejected from NeurIPS 2021.

This does raise questions for program chairs around how selective conferences should be.  The analyses in Sections~\ref{sec:id} and~\ref{sec:ap} suggest that shifting NeurIPS to be significantly more selective may significantly increase the arbitrariness as measured by both the disagreement between committees and the fraction of accepted papers with a different decision upon re-review when compared against appropriate baselines.  However, increasing the acceptance rate may not decrease the arbitrariness appreciably.  We would encourage future program chairs to keep this in mind when considering whether to make the conference more selective.

Some might take this argument to the extreme and suggest that we do away with accept/reject decisions completely, instead relying on readers to judge the quality of papers for themselves, or that NeurIPS moves to a model in which papers are reviewed for technical correctness but not more subjective characteristics like perceived importance of the work, similar to venues like PLOS One. While we find value in the peer review process, we invite the community to debate these options and continue to suggest other ways of improving the peer review process. We hope that the analyses presented here can contribute to this discussion.

\subsection{Limitations}
\label{sec:limitations}

There are two caveats we would like to call out that may impact these results.

First, although we asked authors of duplicated papers to respond to the two sets of reviews independently, there is evidence that some authors put significantly more effort into their responses for the copy that they felt was more likely to be accepted.  In fact, some authors directly informed us that they were only going to spend the time to write a detailed response for the copy of their paper with the higher scores.  Overall, there were 50 pairs of papers where authors only left comments on the copy with the higher average score; interestingly, only two of these papers were ultimately accepted.

To dig into this more, we had 8,765 papers still under review at the time initial reviews were released.  The acceptance rate for the 7,883 papers not in the experiment was 2036/7883 = 25.8\%.  (Note that the overall acceptance rate for the conference was 25.6\%, but this overall rate also includes papers that were withdrawn or rejected for violations of the call for papers prior to initial reviews being released---here we are looking only at papers still under review at this point.)  As discussed above, the average acceptance rate for duplicated papers was 22.7\% (206 papers recommended for acceptance in the original set and 195 papers recommended in the duplicate set, for 401 acceptances out of the total of 882*2 papers).  The 95\% binomial confidence intervals for the two observed rates do not overlap.  Authors changing their behavior may account for this difference.  This confounder may have somewhat skewed the results of the experiment.

Second, when decisions shifted as part of the calibration process, ACs were often asked to edit their meta-reviews to move a paper from ``poster'' to ``reject'' or vice versa, or from ``spotlight'' to ``poster'' or vice versa.  We observed several cases in which ACs made these changes without altering the field for whether a paper ``can be bumped up'' or ``can be bumped down.''  For example, there were nine cases in which it appears that a duplicated paper was initially marked ``poster'' and ``can be bumped down'' and later moved to ``reject,'' ending up marked as the nonsensical ``reject'' and ``can be bumped down.''  This could potentially introduce minor inaccuracies into our analysis of shifted thresholds.

\subsection{Potential Negative Impact on the Community}
\label{sec:negimpact}

Running this experiment required expanding the NeurIPS 2021 program committee's workload by approximately 10\%, which required recruiting additional SACs, ACs, and reviewers. This increases the burden imposed on the machine learning community.  We weighed this cost against the potential benefits in terms of future improvements to the review process that might result as a consequence of this study and hope that on the whole this study benefits the community.  The experiment additionally led to 92 papers being accepted that would not have been otherwise, raising the overall conference acceptance rate from 24.6\% to 25.6\%.  Given the level of noise and inconsistency in the review process that has been demonstrated by this experiment and other work, we do not believe that this meaningfully impacted the overall quality of the conference program.

\section*{Acknowledgments}

We would like to thank the NeurIPS 2021 workflow manager Zhenyu (Sherry) Xue and the entire OpenReview team, especially Melisa Bok, for their support with the experiment.  We would also like to thank everyone who provided feedback on the design of the experiments, especially Corinna Cortes and Neil Lawrence.  Finally, we also thank the reviewers, ACs, and SACs who contributed their time to the review process, and all of the authors who submitted their research to NeurIPS.

Figures~\ref{fig:dis_vs_accept} and \ref{fig:fracchanged_vs_accept} were generated using the XKCD functionality in matplotlib.

\bibliographystyle{plainnat}
\bibliography{refs}

\appendix

\section{NeurIPS Paper Checklist}
\label{app:checklist}

While this report is not a traditional machine learning paper, we have applied the NeurIPS paper checklist to promote responsible research practices. We include our responses below.

\begin{enumerate}

\item For all authors...
\begin{enumerate}
  \item Do the main claims made in the abstract and introduction accurately reflect the paper's contributions and scope?
    \answerYes{}
  \item Did you describe the limitations of your work?
    \answerYes{See Section~\ref{sec:limitations}.}
  \item Did you discuss any potential negative societal impacts of your work?
    \answerYes{See Section~\ref{sec:negimpact}.}
  \item Have you read the ethics review guidelines and ensured that your paper conforms to them?
    \answerYes{}
\end{enumerate}

\item If you are including theoretical results...
\begin{enumerate}
  \item Did you state the full set of assumptions of all theoretical results?
    \answerNA{}
        \item Did you include complete proofs of all theoretical results?
    \answerNA{}
\end{enumerate}

\item If you ran experiments...
\begin{enumerate}
  \item Did you include the code, data, and instructions needed to reproduce the main experimental results (either in the supplemental material or as a URL)?
    \answerNo{As discussed in Section~\ref{sec:methods}, since the data collected included personally identifiable information, to protect authors' privacy, only the 2021 NeurIPS program chairs and workflow manager and OpenReview staff are permitted to access this data. However, for all duplicated papers that were accepted (and any rejected duplicated papers that opted in), both sets of reviews are publicly available on OpenReview.}
  \item Did you specify all the training details (e.g., data splits, hyperparameters, how they were chosen)?
    \answerNA{}
        \item Did you report error bars (e.g., with respect to the random seed after running experiments multiple times)?
    \answerYes{We did not include error bars in Figure~\ref{fig:fracchanged_vs_accept} because there was not a straight-forward way to derive meaningful error bars here.}
        \item Did you include the total amount of compute and the type of resources used (e.g., type of GPUs, internal cluster, or cloud provider)?
    \answerNo{The compute resources required to run our experiments were not significant.}
\end{enumerate}

\item If you are using existing assets (e.g., code, data, models) or curating/releasing new assets...
\begin{enumerate}
  \item If your work uses existing assets, did you cite the creators?
    \answerNA{}
  \item Did you mention the license of the assets?
    \answerNA{}
  \item Did you include any new assets either in the supplemental material or as a URL?
    \answerNA{}
  \item Did you discuss whether and how consent was obtained from people whose data you're using/curating?
    \answerYes{The IRB review determined that explicit consent was not required, but we were asked to notify authors that experiments would be run in the call for papers; see Section~\ref{sec:methods}.}
  \item Did you discuss whether the data you are using/curating contains personally identifiable information or offensive content?
    \answerYes{See Section~\ref{sec:methods}.}
\end{enumerate}

\item If you used crowdsourcing or conducted research with human subjects...
\begin{enumerate}
  \item Did you include the full text of instructions given to participants and screenshots, if applicable?
    \answerYes{See Sections~\ref{app:authoremail} and~\ref{app:forms} in the Appendix.}
  \item Did you describe any potential participant risks, with links to Institutional Review Board (IRB) approvals, if applicable?
    \answerYes{See Section~\ref{sec:methods}.}
  \item Did you include the estimated hourly wage paid to participants and the total amount spent on participant compensation?
    \answerYes{As noted in Section~\ref{sec:background}, program committee members were not compensated for their involvement, nor were authors.}
\end{enumerate}

\end{enumerate}

\section{Review Forms}
\label{app:forms}

The review form used by reviewers contained the following fields (starred fields mandatory):

\begin{itemize}
\item \textbf{Summary$^*$} Briefly summarize the paper and its contributions.
\item \textbf{Main Review$^*$} Provide a full review of the submission, including its originality, quality, clarity, and significance. See \url{https://neurips.cc/Conferences/2021/Reviewer-Guidelines} for guidance on questions to address in your review, and faq for how to incorporate Markdown and LaTeX into your review.
\item \textbf{Limitations And Societal Impact$^*$} Have the authors adequately addressed the limitations and potential negative societal impact of their work? If not, please include constructive suggestions for improvement.
\item \textbf{Ethical Concerns} If there are ethical issues with this paper, please describe them and the extent to which they have been acknowledged or addressed by the authors. See \url{https://neurips.cc/public/EthicsGuidelines} for ethics guidelines.
\item \textbf{Needs Ethics Review$^*$} Should this paper be sent for ethics review? Options:
\begin{itemize}
\item Yes
\item No
\end{itemize}
\item \textbf{Ethics Review Area} If you flagged this paper for ethics review, what area of expertise would it be most useful for the ethics reviewer to have? Please click all that apply. Options: 
\begin{itemize}
\item Discrimination / Bias / Fairness Concerns
\item Inadequate Data and Algorithm Evaluation
\item Inappropriate Potential Applications \& Impact  (e.g., human rights concerns)
\item Privacy and Security (e.g., consent)
\item Legal Compliance (e.g., GDPR, copyright, terms of use)
\item Research Integrity Issues (e.g., plagiarism)
\item Responsible Research Practice (e.g., IRB, documentation, research ethics)
 \item I don’t know
\end{itemize}
\item \textbf{Time Spent Reviewing$^*$} How much time did you spend reviewing this paper (in hours)?
\item \textbf{Rating$^*$}  Please provide an ``overall score'' for this submission. Options: 
\begin{itemize}
\item 10: Top 5\% of accepted NeurIPS papers, seminal paper
\item 9: Top 15\% of accepted NeurIPS papers, strong accept
\item 8: Top 50\% of accepted NeurIPS papers, clear accept
\item 7: Good paper, accept
\item 6: Marginally above the acceptance threshold
\item 5: Marginally below the acceptance threshold
\item 4: Ok but not good enough - rejection
\item 3: Clear rejection
\item 2: Strong rejection
\item 1: Trivial or wrong
\end{itemize}
\item \textbf{Confidence$^*$} Please provide a ``confidence score'' for your assessment of this submission. Options:
\begin{itemize}
\item 5: You are absolutely certain about your assessment. You are very familiar with the related work and checked the math/other details carefully.
\item 4: You are confident in your assessment, but not absolutely certain. It is unlikely, but not impossible, that you did not understand some parts of the submission or that you are unfamiliar with some pieces of related work.
\item 3: You are fairly confident in your assessment. It is possible that you did not understand some parts of the submission or that you are unfamiliar with some pieces of related work. Math/other details were not carefully checked.
\item 2: You are willing to defend your assessment, but it is quite likely that you did not understand central parts of the submission or that you are unfamiliar with some pieces of related work. Math/other details were not carefully checked.
\item 1: Your assessment is an educated guess. The submission is not in your area or the submission was difficult to understand. Math/other details were not carefully checked.
\end{itemize}
\item \textbf{Code of Conduct$^*$} (Checkbox) While performing my duties as a reviewer (including writing reviews and participating in discussions), I have and will continue to abide by the NeurIPS code of conduct.
\end{itemize}

The meta-review form used by ACs contained the following fields (starred fields mandatory):

\begin{itemize}
\item \textbf{Recommendation$^*$} Please recommend a decision for this submission. Options:
\begin{itemize}
\item Accept (Oral)
\item Accept (Spotlight)
\item Accept (Poster)
\item Reject 
\end{itemize}
\item \textbf{Confidence$^*$} Please qualify your recommendation. (Your response will not be shared with authors or reviewers.) Options:
\begin{itemize}
\item You are absolutely certain.
\item This decision can be bumped up (e.g., you are recommending to accept this paper as a poster but it would also make a fine spotlight).
\item This decision can be bumped down (e.g., you have a preference for accepting this paper as a poster but you wouldn’t mind if it is rejected).
\item You are not certain what the right decision should be.  This is a submission you wish to discuss further with the SAC.
\end{itemize}
\item \textbf{Metareview$^*$} Please provide a meta-review for this submission. Your meta-review should explain your decision to the authors. Your comments should augment the reviews, and explain how the reviews, author response, and discussion were used to arrive at your decision. If you want to make a decision that is not clearly supported by the reviews, perhaps because the reviewers did not come to a consensus, please justify your decision appropriately.
\item \textbf{Review History$^*$} Have you previously reviewed or area chaired (a version of) this work for another archival venue? (This information is only requested in order to collect aggregate statistics for NeurIPS 2021. Your response will not be shared with authors or reviewers, nor will it affect the submission's chance of acceptance.) 
\item \textbf{Consider For An Award} (Checkbox)  Yes, this paper should be seriously considered for an award.
\end{itemize}

\section{Disclosure to Authors}
\label{app:authoremail}

The following is the text of the email that was sent to authors whose papers were duplicated before initial reviews were released.

\begin{quote}
Subject: NeurIPS reviews: you are part of an experiment, please read this email in full

Dear \textless{firstname}\textgreater,

Preliminary NeurIPS reviews will be released tomorrow and the author response period will begin.  When this happens, you will receive two independent sets of reviews for your paper \textless{paper title}\textgreater.  This email explains why this will happen and how you should respond to these reviews.  We ask that you do not discuss the information in this email with anyone outside of the co-authors on your affected submission(s).

NeurIPS has a long history of experimentation.  For many years, Program Chairs have run randomized controlled experiments to measure the quality of the review process and the effectiveness of proposed alternatives. In 2014, NeurIPS ran an experiment in which 10\% of submissions were reviewed by two independent program committees to quantify the randomness in the review process.  This year we are repeating a variant of this experiment to see how the quality of the review process has changed over time.  Your paper is among the 10\% randomly chosen to receive two independent sets of reviews.

Below we answer some questions you may have as you prepare for the author response period.

Q: Can I discuss the experiment with colleagues or post about it on social media? \\
A: No. The experiment will not be publicly announced until the review process is over.  At this time, we are informing only authors of papers in the replicated 10\% so that you know how to respond to reviews.  We ask that you please keep the experiment and the fact that you have received two sets of reviews confidential.

Q: How should I respond to my reviews? \\
A: You should respond to each set of reviews independently.  The reviewers, area chairs, and senior area chairs assigned to each copy of your paper do not know that there is more than one copy of the paper under review.

Q: What about the rolling discussion? \\
A: The same holds for the rolling discussion between reviewers and authors that will follow the initial author response period.  Respond to the two discussions independently.

Q: How will a final decision be made about my paper? \\
A: If both sets of reviewers make the same accept/reject recommendation, this recommendation will be followed.  If a single set recommends acceptance, the paper will be accepted as long as the other set does not identify a fatal flaw (e.g., an error in a major result that cannot easily be fixed).  The Program Chairs will make the final call in such cases.

Q: Will both sets of reviews be publicly released if my paper is accepted or if I opt in to have my rejected paper be public? \\
A: Yes, both sets of reviews will be released (along with your responses to the reviews and any follow up discussion with the program committee assigned to your paper).

Q: Who should I contact if I have questions? \\
A: Please contact the Program Chairs directly at neurips2021pcs@gmail.com.

Alina Beygelzimer, Yann Dauphin, Percy Liang, and Jenn Wortman Vaughan \\
NeurIPS 2021 Program Chairs
\end{quote}

\end{document}